\documentclass[runningheads]{llncs}
\usepackage{wrapfig}
\usepackage{multirow}
\usepackage{graphicx}
\usepackage{amsfonts}
\usepackage{mathtools}
\usepackage{amsmath}
\usepackage{amssymb}
\usepackage{cite}
\usepackage{footnote}
\usepackage{array}
\usepackage{subfigure}
\usepackage{setspace}
\usepackage{booktabs}
\usepackage[misc,geometry]{ifsym} 
\usepackage[ruled,vlined]{algorithm2e}
\usepackage{ulem}

\usepackage[colorlinks,
            linkcolor=green,
            anchorcolor=red,
            citecolor=blue
            ]{hyperref}

\begin{document}
\title{Leveraging Undiagnosed Data for Glaucoma Classification with Teacher-Student Learning}
\titlerunning{Leveraging Undiagnosed Data for Glaucoma Classification with TSL}

\author{Junde Wu \inst{1,2} \and
Shuang Yu \inst{1}\textsuperscript{(\Letter)}  \and 
Wenting Chen \inst{1} \and
Kai Ma \inst{1} \and
Rao Fu \inst{2} \and
Hanruo Liu \inst{3} \and
Xiaoguang Di \inst{2}  \and
Yefeng Zheng \inst{1}
}

\authorrunning{J. Wu et al.}

\institute{	
Tencent Healthcare, Tencent, Shenzhen, China \\
\email{shirlyyu@tencent.com} \\
\and
 Control and Simulation Center, Harbin Institute of Technology, Harbin, China\\
\and
Beijing Tongren Hospital, Capital Medical University, Beijing, China \\
}

\maketitle             
\begin{abstract}

Recently, deep learning has been adopted to the glaucoma classification task with performance comparable to that of human experts. However, a well trained deep learning model demands a large quantity of properly labeled data, which is relatively expensive since the accurate labeling of glaucoma requires years of specialist training. In order to alleviate this problem, we propose a glaucoma classification framework which takes advantage of not only the properly labeled images, but also undiagnosed images without glaucoma labels. 
To be more specific, the proposed framework is adapted from the teacher-student-learning paradigm. The teacher model encodes the wrapped information of undiagnosed images to a latent feature space, meanwhile the student model learns from the teacher through knowledge transfer to improve the glaucoma classification.
For the model training procedure, we propose a novel training strategy that simulates the real-world teaching practice named as ``Learning To Teach with Knowledge Transfer (L2T-KT)", and establish a``Quiz Pool" as the teacher's optimization target. Experiments show that the proposed framework is able to utilize the undiagnosed data effectively
to improve the glaucoma prediction performance.  


\end{abstract}

\section{Introduction}

Glaucoma is the leading cause of irreversible vision loss that primarily damages the optic cup/disc and surrounding optic nerve\cite{tham2014global}. Recently, deep learning methods have achieved rapid advancement and been widely adopted to the automatic glaucoma classification using fundus images \cite{LAG_5,LAG_22,LAG_23,LAG}. However, a substantially large amount of properly labeled data is generally required for training deep learning models, which might not be easily accessed as the accurate grading of glaucoma, especially at the early stage, requires years of expertise for glaucoma specialists.

Beyond the publicly available glaucoma classification datasets, on the other hand, there have been several high-quality publicly available datasets for cup/disc segmentation, but without image-level glaucoma labels\cite{seg_data, seg_data2, seg_data3}.
The Cup-to-Disc Ratio (CDR) parameter, which can be easily computed from the cup/disc masks, is one of the most important clinical parameters for the diagnosis of glaucoma. Generally, patients with a CDR value higher than 0.6 are considered as glaucoma suspects and a higher CDR value indicates a higher probability of having glaucoma \cite{garway1998vertical}. 
This inspires us to take advantage of the images with only cup/disc segmentation masks to improve the glaucoma classification performance.

In order to properly utilize the undiagnosed images, we propose to transfer the knowledge of pixel-wise cup/disc labels to the learner model via a teacher-student learning paradigm, which has been a popular and effective way to incorporate any prior information. However, most existing teacher-student learning methods learn a compact student from a stronger but more complex teacher, for the purpose of knowledge distillation\cite{born_again1,born_again2,born_again3}. Some other methods\cite{l2t,l2t-dl} learn a teacher to improve the student's training speed or performance, but they assume that the ground-truth labels of all the training samples are available, which is not true under our scenario. To the best of our knowledge, there is still a research gap of how to utilize the teacher model to learn from undiagnosed data to improve the student model's performance on glaucoma classification.

In this paper, we aim to address this research gap by proposing a novel training strategy, named ``Learning To Teach with Knowledge Transfer (L2T-KT)" with a reserved quiz pool, imitating the real-world teaching practice. In L2T-KT, the teacher learns to encode the undiagnosed fundus images and the corresponding cup/disc masks to a latent feature space with the ultimate goal of improving the student's performance on the quiz pool. Meanwhile, the student is updated by the supervision of the teacher through knowledge transfer. 
Three major contributions are made with this paper. Firstly, we propose to adapt the teacher-student learning paradigm to the glaucoma screening task and verify the feasibility to utilize undiagnosed images to improve the glaucoma classification performance. 
Secondly, we propose a novel training strategy of L2T-KT and quiz pool to update the teacher model with undiagnosed images, which enables the teacher to extract potential important features from the undiagnosed images and further improve the performance of the student model via knowledge transfer. Finally, the proposed method can be easily extended to learn from totally unlabeled images or transductive learning to improve the model performance.

\section{Methodology}

Consider all the collected fundus images as dataset $\mathcal{D}$ that can be divided into the primary training set $\mathcal{D}_{d}$ with glaucoma classification labels, and auxiliary training set $\mathcal{D}_{m}$ with cup/disc masks but without glaucoma labels. 
The primary training set is denoted as: $(x_{1}^{d},y_{1}),(x_{2}^{d},y_{2})...(x_{n}^{d},y_{n})\in \mathcal{D}_{d}$, where $x_{i}^{d}$ denotes the fundus image, $y_{i} \in \{0, 1\}$ denotes the glaucoma label for the input image $x_{i}^{d}$. 
Meanwhile, the auxiliary training data is denoted as $(x_{1}^{m},m_{1}),(x_{2}^{m},m_{2})...(x_{c}^{m},m_{c})\in \mathcal{D}_{m}$, where $x_{i}^{m}$ is the fundus image, $m_{i}$ denotes the optic cup/disc mask.
Furthermore, the primary training dataset is further divided into textbook pool $\mathcal{D}_{TP}$ (for training the student model) and quiz pool $\mathcal{D}_{QP}$ (for updating the teacher model).
Provided with these datasets, the target of this research is to construct a framework that can learn a mapping function $f(x_{i}^{d}) = y_{i}$ using both primary dataset $\mathcal{D}_{d}$ and auxiliary dataset $\mathcal{D}_{m}$, and thus is expected to outperform the mapping function $g(x_{i}^{d}) = y_{i}$ that learns only using the primary training dataset. 
Since the teacher-student learning paradigm has been widely used for knowledge distillation and proved effective for extracting latent information, we construct a deep learning model based on teacher-student learning for the glaucoma screening task.

\begin{figure}[t]

	\centering
	\includegraphics[width = \textwidth]{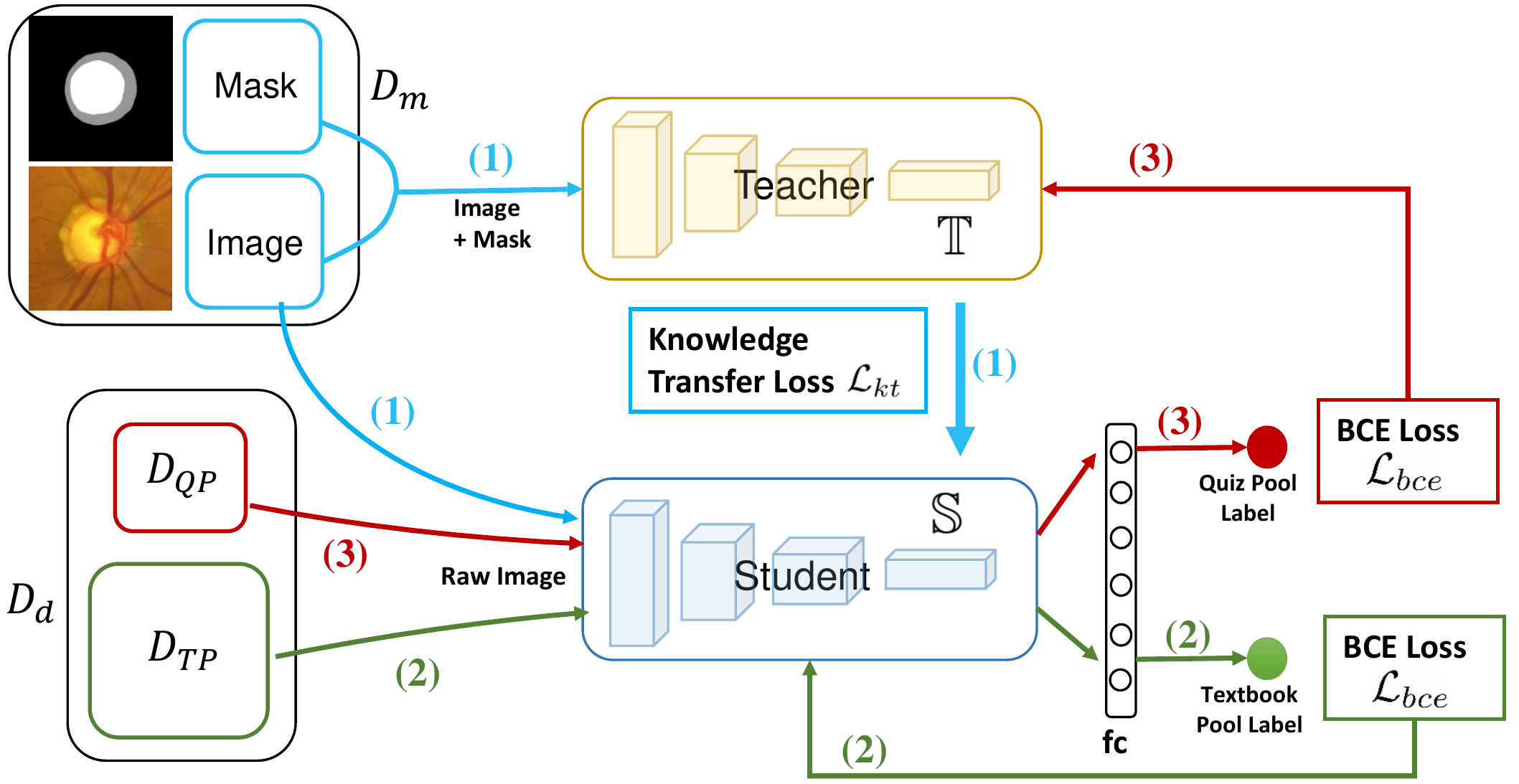}

	\caption{Framework and data flow of the proposed L2T-KT framework. Stage (1), train the student model with knowledge transfer loss; Stage (2), update the student model with textbook pool data using binary cross entropy loss;  Stage (3), train the teacher model with quiz pool data using binary cross entropy loss.}
	\label{fig:overallmodel}

\end{figure}

As shown in Fig. \ref{fig:overallmodel}, the overall framework contains two networks: the teacher model $f_{t}$ and the student model $f_{s}$.
The teacher model is a convolutional neural network which encodes the fundus images together with the corresponding cup/disc masks into a latent feature space. And the student model shares the same feature extraction backbone as that of the teacher. In this paper, the state-of-the-art classification network EfficientNet (B4) is adopted as the feature extraction backbone \cite{effecientnet}. Different from the teacher model, the student model contains a fully connected layer (fc) to make predictions for glaucoma.

The proposed framework is optimized via an iterative three-stage training strategy. In the first stage, the student model is supervised by the teacher model using data from the auxiliary dataset $\mathcal{D}_{m}$ with knowledge transfer loss, as marked by the cyan color data flow in Fig. \ref{fig:overallmodel}. 
In the second stage, the student model is further optimized by the ground-truth glaucoma labels from the textbook pool $D_{TP}$ with binary cross entropy (BCE) loss, as marked by the green data flow.
In the last stage, as marked by the red data flow, the teacher model is updated with the proposed `learning to teach' strategy using the quiz pool $D_{QP}$. 
The detailed updating strategy of the framework is explained in Algorithm \ref{algorithm:overall}. Note that the input to the student model is only the fundus image, meanwhile input to the teacher model contains both image and its corresponding cup/disc mask from the auxiliary dataset $\mathcal{D}_{m}$.

\begin{algorithm}
	
	\label{algorithm:overall}
	\SetAlgoLined
	Given networks: teacher model $f_{t}$, student model $f_{s}$; \\
	Datasets: primary training dataset $D_{d}$, auxiliary training dataset $D_{m}$ \;
	Initialize textbook pool $D_{TP}$ and quiz pool $D_{QP}$ by randomly split $D_{d}$\;
	Initialize $f_{s}$ randomly and initialize $f_{t}$ with pretrained baseline parameters\;
	\While{Training}{
				
		Sample $(x,m)$ from $\mathcal{D}_{m}$ \;
		Send $(x,m)$ to $f_{t}$ to get $\mathbb{T}$\;
		Send $x$ to $f_{s}$ to get $\mathbb{S}$\;
		Update $f_{s}$ with knowledge transfer loss $\mathcal{L}_{kt}$ by Eqn. \ref{equation:s_loss_kt}\;
		
		Sample $(x, y)$ from $D_{TP}$\;
		Update $f_{s}$ with BCE loss by Eqn. \ref{equation:s_loss_bce}\;

		Sample $(x, y)$ from $D_{QP}$\;
		Update $f_{t}$ through L2T-KT by Eqn. \ref{equation:L2T-1} and Eqn.\ref{equation:L2T-2}\;

		Update $D_{QP}$ by Eqn. \ref{equation:dqp_ds}, Eqn. \ref{equation:dqp_p}\;
	}
	\caption{Overall learning process of the proposed model}
	
\end{algorithm}

\subsection{Quiz Pool}\label{sec:quiz-pool}

In the real-world teaching practice, students have access to the textbook content, but have no access to the answers of the quiz problems, which are used by the teacher to evaluate the student's performance and update the teaching strategy based on the evaluation scores.
Inspired by this scenario, we propose to split the primary data set $\mathcal{D}_{d}$ into two subsets, the textbook pool $\mathcal{D}_{TP}$ and the quiz pool $\mathcal{D}_{QP}$. The student model learns with the ground-truth glaucoma labels from the textbook pool $\mathcal{D}_{TP}$; meanwhile, the student's performance is evaluated on the quiz pool $\mathcal{D}_{QP}$. The evaluation score is used to update the teacher model. 

In this paper, $\mathcal{D}_{TP}$ and $\mathcal{D}_{QP}$ are split with two different approaches. The first method is the static quiz pool, where $\mathcal{D}_{QP}$ is randomly selected from the auxiliary set $\mathcal{D}_{d}$ and kept the same during the training procedure. In this paper, 20\% of samples are randomly selected from $\mathcal{D}_{d}$ as the quiz pool $\mathcal{D}_{QP}$.

In the second method, we propose to update the quiz pool dynamically during the training process, i.e., the dynamic quiz pool. 
Practically, teachers often reserve the important or difficult contents for the quiz problems. Similar to this idea, a dynamic quiz pool is established, which focuses on the difficult cases and positive cases, since missing the positive cases is at higher risk for glaucoma screening. The pool is dynamically updated depending on the samples' difficulty reported by comparing the student's predictions with the glaucoma labels.
The difficulty of individual samples reported by the student can be obtained with:
\begin{equation}\label{equation:dqp_ds}
\psi^{s} =  (1-y)y' + y(1-y'),
\end{equation}
where $y \in \{0,1\}$ denotes the ground-truth glaucoma label and $y'$ denotes the student's prediction. Then, the probability of a sample from $D_{d}$ being selected into the dynamic quiz pool is calculated by:
\begin{equation}
\label{equation:dqp_p}
P_{d} = \left( \alpha (1-y) + y \right) \cdot (\psi^{s})^{\gamma} \cdot
\left( 1+e^{-\sigma[(1-\bar{\psi}^{q})-\mu]}\right) ^{-1},
\end{equation}
where $\alpha$ denotes the relative importance of negative samples compared to that of positive samples,
and $(\psi^{s})^{\gamma}$ encourages the pool to focus on the difficult samples. In this paper, $\alpha$ and 
$\gamma$ are empirically set to 0.7 and 2, respectively. The last term is a shifted sigmoid function that controls the average difficulty $\bar{\psi}^{q}$ of the quiz pool within a reasonable range. It encourages the quiz pool to retain the test content if it is challenging to the student, while changing the pool if it is too easy. In this work, $\sigma$ and $\mu$ are set to 16 and 0.5, respectively. In addition, when a new sample is added to the quiz pool, the easiest sample will be dropped, so as to keep a constant size of the quiz pool. In the implementation, the quiz pool is updated every epoch.

\subsection{Student Update Through Knowledge Transfer}\label{sec:student}

The student model is trained on both textbook pool $\mathcal{D}_{TP}$ of the primary dataset with glaucoma label and the auxiliary dataset $\mathcal{D}_m$ without glaucoma label. 

In the first stage, the student model is trained with the auxiliary set $\mathcal{D}_m$ and supervised by the teacher model with knowledge transfer. More specifically, the color fundus images are first concatenated with the corresponding cup/disc masks and then fed to the teacher model, which will encode the input to the latent feature maps $\mathbb{T}$.
Meanwhile, the same set of color fundus images (without masks) will be sent to the student model as well to get the latent feature maps $\mathbb{S}$. Then the knowledge transfer (KT) loss between $\mathbb{T}$ and $\mathbb{S}$ can be computed by learning the domain-invariant latent representations with Centered Kernel Alignment (CKA) \cite{cka}, as below:

\begin{equation}\label{equation:s_loss_kt}
\mathcal{L}_{kt} = \mathrm{CKA} \, (\mathbb{T} \, \mathbb{T}^{T}, \mathbb{S} \, \mathbb{S}^{T}) = \frac{\lVert \mathbb{S}^{T} \, \mathbb{T} \rVert^{2}_{F}}{\lVert \mathbb{T}^{T} \, \mathbb{T} \rVert_{F} \, \lVert \mathbb{S}^{T} \, \mathbb{S} \rVert_{F}},
\end{equation}
where $\lVert \cdot \rVert_{F}$ denotes the Frobenius norm.

In the second stage, the student model is further trained with data from the textbook pool $\mathcal{D}_{TP}$, which contains fundus images and the ground-truth glaucoma labels. At this stage, the student model is directly supervised by the ground-truth glaucoma labels with binary cross entropy (BCE) loss:
\begin{equation}\label{equation:s_loss_bce}
\mathcal{L}_{bce} =- \left(  y \, \log y' +(1-y) \, \log (1-y') \right) ,
\end{equation}
where $y$ is the ground-truth glaucoma label, and $y'$ is the student's prediction.

\subsection{Teacher Update Trough L2T-KT}\label{sec:teacher-learning}

Following the real-world teaching practice where teachers often update their teaching strategies based on students' feedback, we propose to update the teacher model parameters based on the student's performance on the constructed quiz pool $D_{QP}$.
Formally speaking, consider a teacher network with parameters $\theta_{t}$ as $f_{\theta_{t}}$ and a student network with parameters $\theta_{s}$ as $f_{\theta_{s}}$. The response of $f_{\theta_{t}}$ to a concatenated fundus image and mask $(x,m) \in \mathcal{D}_{m}$ is $f_{\theta_{t}}(x,m) $. The response of $f_{\theta_{s}}$ to the raw fundus image $x \in (x,m)$ is $f_{\theta_{s}}(x)$. The first step of L2T-KT training strategy is computing the update of the student parameters $\theta_{s}$ with the knowledge transfer loss between $f_{\theta_t}(x,m)$ and $f_{\theta_{s}}(x)$, which can be expressed in the gradient descent format as:
\begin{equation}\label{equation:L2T-1}
\hat{\theta_{s}} = \theta_{s} - \lambda_{s}  \frac{\partial \mathcal{L}_{kt}[f_{\theta_t}(x,m),f_{\theta_{s}}(x)]}{\partial \theta_{s}}, 
\end{equation}
where $\lambda_{s}$ denotes the learning rate of the student model and is set as $3\times10^{-4}$. After the knowledge transfer and student parameter update, we denote the refreshed student as $f_{\hat{\theta_{s}}}$.

The teacher's goal is learning to teach the student to achieve better performance on the quiz. In other words, the optimization target of the teacher is the refreshed student $f_{\hat{\theta_{s}}}$ to perform better than $f_{\theta_{s}}$ on the same $\mathcal{D}_{QP}$. 
Let $f_{\hat{\theta_{s}}}(q)$ denotes the prediction of $f_{\hat{\theta_{s}}}$ over a random sample $q \in \mathcal{D}_{QP}$ and $y_{q}$ denotes the ground-truth label of $q$, the teacher can be optimized by minimizing the BCE loss between $f_{\hat{\theta_{s}}}(q)$ and $y_{q}$ with gradient descent. It is theoretically feasible because as shown in Eqn. \ref{equation:L2T-1}, teacher parameter $\theta_{t}$ is a variable of the updated student's parameter $\hat{\theta_{s}}$. 
Therefore, the partial derivative of $f_{\hat{\theta_{s}}}(q)$ w.r.t $\theta_{t}$ can be computed, and $\theta_{t}$ can be updated via:
\begin{equation}\label{equation:L2T-2}
\hat{\theta_{t}} = \theta_{t} - \lambda_{t}  \frac{\partial \mathcal{L}_{bce}[f_{\hat{\theta_{s}}}(q),y_{q}]}{\partial \theta_{t}}, 
\end{equation}  
where $\lambda_{t}$ is the learning rate of the teacher model and set as $3\times10^{-4}$. 
We compute the partial derivative of $f_{\hat{\theta_{s}}}(q)$ w.r.t the teacher parameter $\theta_{t}$, rather than its own parameter $\hat{\theta_{s}}$ as commonly used. That is because we aim at making the teacher to learn how to teach a better student, but not making the student to learn by itself. Note that the refreshed student is only temporarily used in L2T-KT, which will not change the parameters of the original student.

\section{Experiments}

\subsubsection{Datasets}

The data utilized in this work mainly originates from two sources: the primary dataset $\mathcal{D}_{d}$ with glaucoma labels from Beijing Tongren Hospital with approval obtained from the institutional review board, and the auxiliary dataset $\mathcal{D}_{m}$ with cup/disc segmentation masks from publicly available dataset RIGA \cite{seg_data}. The primary dataset contains in total of 3,830 fundus images graded by certified glaucoma specialists, including 1,586 glaucoma and 2,244 non-glaucoma images. We randomly selected 60\% images as the training set, 15\% as the validation set and the rest 25\% as the test set to evaluate the model performance. The
RIGA dataset contains 650 fundus images with pixel-wise cup/disc masks labeled by experts, but image-level glaucoma labels are not provided \cite{seg_data}.

\subsection{Ablation Studies}

Ablation studies have been conducted to evaluate the effectiveness of the proposed framework under different setups of the quiz pool, including the the static quiz pool and dynamic quiz pool. 
The comparison baseline method utilizes the same backbone as that of the teacher/student model, i.e., EfficientNet-B4, and trained with the glaucoma/non-glaucoma labels of the primary training set. Four metrics are adopted to evaluate the model performance, including accuracy (Acc), sensitivity (Sen), specificity (Spec) and area under the receiver operating characteristic curve (AUC).

Table \ref{table:ablation} shows the quantitative comparisons of the baseline method and the proposed framework trained on undiagnosed auxiliary dataset under different settings of the quiz pool. Compared with the baseline method using purely labeled data, training using both labeled and undiagnosed data with the proposed L2T-KT framework with a static quiz pool increases the AUC with 2.39\% and accuracy with 2.33\%. In addition, by changing the static quiz pool to a dynamic quiz pool, the model performance is further improved, with an obvious improvement on the model sensitivity, since the dynamic quiz pool favors the positive cases. Clinically, for the glaucoma screening task, a higher sensitivity measure is much more important than specificity, so as not to miss the potential glaucoma patients.

\setlength{\tabcolsep}{3pt}
\renewcommand{\arraystretch}{1.1}
\begin{table}[!t]
	\caption{Performance comparison (\%) of leveraging undiagnosed data under different settings of the quiz pool.}
	\centering
	\begin{tabular}{c|p{1.2cm}<{\centering}|p{1.3cm}<{\centering}|p{1.2cm}<{\centering}|p{1.2cm}<{\centering}|p{1.2cm}<{\centering}|p{1.2cm}<{\centering}}
		\hline
		& Static & Dynamic & Acc    & Sen    & Spec   & AUC    \\ \hline
	    Baseline$^*$ &  &   & 90.69 & 90.10 & 93.78 & 95.77 \\ \hline
		 Proposed$^\dagger$ & $\surd$   &     & 93.02 & 90.70 & \textbf{94.53} & 98.16 \\ \hline
		     Proposed$^\ddagger$ & &   $\surd$  & \textbf{93.29} & \textbf{96.03} & 91.42 & \textbf{98.29} \\ \hline

    \multicolumn{7}{l}{\footnotesize{$*$: Baseline method using purely labeled data;}} \\
	\multicolumn{7}{l}{\footnotesize{$\dagger$: Proposed method: labeled data + undiagnosed data + static quiz pool;}} \\
	\multicolumn{7}{l}{\footnotesize{$\ddagger$: Proposed method: labeled data + undiagnosed data+dynamic quiz pool;}} \\
	
	\end{tabular}
	\label{table:ablation}
\end{table}

\subsection{Auxiliary Data Setting}

We have also evaluated the model performance under different settings of the auxiliary dataset. The evaluation is conducted on both the private dataset and a publicly available glaucoma classification dataset LAG \cite{LAG}, which contains
1,711 glaucoma images and 3,143 non-glaucoma images. Apart from the undiagnosed auxiliary set with ground-truth cup/disc masks (RIGA), we have also alternatively trained on the totally unlabeled auxiliary set, by producing pseudo masks of RIGA images using a state-of-the-art cup/disc segmentation algorithm \cite{wang2019patch}. 
As Table \ref{tabel:comparision} lists, by using an auxiliary set with pseudo cup/disc masks, the model performance degenerates slightly compared with the standard undiagonosed auxiliary set using ground-truth cup/disc masks, with the AUC value drop of 0.17$\%$ and 0.45$\%$ for LAG and private set, respectively. However, compared with the baseline model in Table \ref{table:ablation}, the proposed framework using pseudo labels still surpasses that of the baseline model with a remarkable margin, with an AUC improvement of 2.07$\%$ for the private set. 

The performance of the proposed method can be further improved when it is authorized to get access to the raw images of the test dataset, i.e., in the transductive learning scenario. When taking the raw fundus images of the test set and their pseudo cup/disc masks as the auxiliary data (denoted as `Transductive' in Table \ref{tabel:comparision}), the proposed method achieves the best performance, with an AUC score of 99.51$\%$ and 98.41$\%$ for the LAG and private set, respectively. This indicates the expandability and effectiveness of the proposed L2T-KT framework.

\subsection{Comparing with State-of-the-Art}

\renewcommand{\arraystretch}{1.1} 
\begin{table}[!t]
	\centering
	\caption{Performance comparison (\%) with other methods.}
	\resizebox{\columnwidth}{!}{%
	\begin{tabular}{c|cccc|cccc}
		\hline
		& \multicolumn{4}{c|}{LAG} & \multicolumn{4}{c}{Private} 
		\\ \hline
		& Acc & Sen & Spec & AUC & Acc & Sen & Spec & AUC\\ \hline
		Li et al. \cite{LAG} (supervised) & 95.3 & 95.4 & \textbf{95.2} & 97.5  & - & - & - & - \\ 
		Fu et al. \cite{fu2018disc} (supervised) & 93.88 & 96.79 & 92.29 & 98.27 & 91.94 & 91.38 & 92.30 & 96.70  \\ 
		Pinto et al. \cite{gan} (semi) & 92.75 & 92.30 & 93.16 & 97.11  & 91.85 & 92.51 & 89.79 & 96.12 \\ 
		Pinto et al.\cite{gan} (trans) & 93.77 & 93.26 & 94.68 & 97.95 & 92.03 & 92.75 & \textbf{92.60} & 97.45 \\ 
		Ghamdi et al.\cite{pseudo_retrain} (semi) & 94.11 & 97.43 & 92.29 & 98.16  & 91.76 & 93.19 & 90.83 & 96.88 \\ 
		Ghamdi et al.\cite{pseudo_retrain} (trans) & 95.01 & 97.75 & 93.52 & 98.73  & 92.57 & 94.10 & 91.57 & 97.39 \\ \hline
		Auxiliary-GT$^*$ (proposed) & 95.81 & 98.40 & 94.22 & 99.49  & 93.29 & 96.03 & 91.42 & 98.29 \\ 
		Auxiliary-Psd$^\dagger$ (proposed) & 95.47 & \textbf{98.72} & 93.70 & 99.32  & 92.84 & 95.01 & 91.42 & 97.84 \\ 
		Transductive$^\ddagger$ (proposed) & \textbf{96.04} & \textbf{98.72} & 94.75 & \textbf{99.51}  & \textbf{93.64} & \textbf{96.37} & 91.82 & \textbf{98.41} \\ \hline
		
	\multicolumn{9}{l}{\footnotesize{$*$: RIGA dataset with ground-truth cup/disc mask is used as undiagnosed auxiliary set;}} \\
	\multicolumn{9}{l}{\footnotesize{$\dagger$: RIGA dataset with pseudo cup/disc mask is used as totally unlabeled auxiliary set;}} \\
	\multicolumn{9}{l}{\footnotesize{$\ddagger$: Test set with pseudo cup/disc mask is used as totally unlabeled auxiliary set;}} \\
	
	\end{tabular}
	}

	\label{tabel:comparision}
\end{table}

The proposed framework has been compared with other state-of-the-art methods on the glaucoma classification task, including two fully-supervised methods\cite{fu2018disc,LAG} and two semi-supervised methods\cite{pseudo_retrain,gan}. The semi-supervised methods take RIGA or the test set as the unlabeled datasets, which are denoted as `semi' and `trans', respectively. 
As listed in Table \ref{tabel:comparision}, the proposed method achieves the best performance on both the private and LAG datasets, especially for the sensitivity metric, indicating the effectiveness of the proposed method in exploiting extra undiagnosed data.
In addition, different from the existing methods of designing complex architectures or using model ensembling, the proposed method adopts a plug-and-play training strategy without any change on the backbone network. At the prediction stage, only the student network is used, ensuring fast computational speed during inference.

\section{Conclusion}

Many publicly available datasets contain images with cup/disc masks but without image-level glaucoma labels. In order to fully exploit those undiagnosed
images for the glaucoma screening task, we proposed a novel training strategy Learning to Teach with Knowledge Transfer (L2T-KT), which enabled the model
to learn from those undiagnosed images through teacher-student paradigm. Detailed experiments revealed that the proposed method could not only improve the glaucoma screening performance through learning from the data with cup/disc masks, but also could be easily extended to learn from the completely unlabeled data with pseudo labels and improved the test set performance via transductive learning. Future works will continue to explore the potential of the proposed method and optimize the time efficiency at the training stage.

\subsubsection{Acknowledgment}
This work was funded by the Key Area Research and Development Program of Guangdong Province, China (No. 2018B010111001), National Key Research and Development Project (No. 2018YFC2000702) and Science and Technology Program of Shenzhen, China (No. ZDSYS201802021814180).

\bibliographystyle{splncs04}
\bibliography{paper1013}

\end{document}